\documentclass[10pt, a4paper]{article}

\usepackage[]{lrec-coling2024}

\usepackage{microtype}
\usepackage{graphicx}
\usepackage{subfigure}
\usepackage{booktabs}
\usepackage{amsmath}
\usepackage{multirow}
\usepackage{CJKutf8}

\title{A Multimodal In-Context Tuning Approach for
E-Commerce Product Description Generation}

\name{Yunxin Li$^{1}$, Baotian Hu$^{1 *}$\thanks{* Corresponding author.}, Wenhan Luo$^{2}$, Lin Ma$^{3}$, Yuxin Ding$^{1}$, and Min Zhang$^{1}$} 

\address{$^{1}$ Harbin Institute of Technology, Shenzhen, China \\
        $^{2}$ Hong Kong University of Science and Technology, Hong Kong \\
         $^{3}$ Meituan, Beijing, China \\
         liyunxin987@163.com, hubaotian@hit.edu.cn, whluo.china@gmail.com\\
         }

\abstract{
In this paper, we propose a new setting for generating product descriptions from images, augmented by marketing keywords. It leverages the combined power of visual and textual information to create descriptions that are more tailored to the unique features of products. For this setting, previous methods utilize visual and textual encoders to encode the image and keywords and employ a language model-based decoder to generate the product description. However, the generated description is often inaccurate and generic since same-category products have similar copy-writings, and optimizing the overall framework on large-scale samples makes models concentrate on common words yet ignore the product features. To alleviate the issue, we present a simple and effective \textbf{M}ultim\textbf{od}al \textbf{I}n-\textbf{C}ontext \textbf{T}uning approach, named \textbf{ModICT}, which introduces the similar product sample as the reference and utilizes the in-context learning capability of language models to produce the description. During training, we keep the visual encoder and language model frozen, focusing on optimizing the modules responsible for creating multimodal in-context references and dynamic prompts. This approach preserves the language generation prowess of large language models (LLMs), facilitating a substantial increase in description diversity. To assess the effectiveness of ModICT across various language model scales and types, we collect data from three distinct product categories within the E-commerce domain. Extensive experiments demonstrate that ModICT significantly improves the accuracy (by up to 3.3\% on Rouge-L) and diversity (by up to 9.4\% on D-5) of generated results compared to conventional methods. Our findings underscore the potential of ModICT as a valuable tool for enhancing the automatic generation of product descriptions in a wide range of applications. Data and code is at \url{https://github.com/HITsz-TMG/Multimodal-In-Context-Tuning}.
 \\ \newline  }

\begin{document}

\maketitleabstract

\section{Introduction}

\begin{figure}[t]
    \centering
    \includegraphics[width=0.45\textwidth,height=0.35\textwidth]{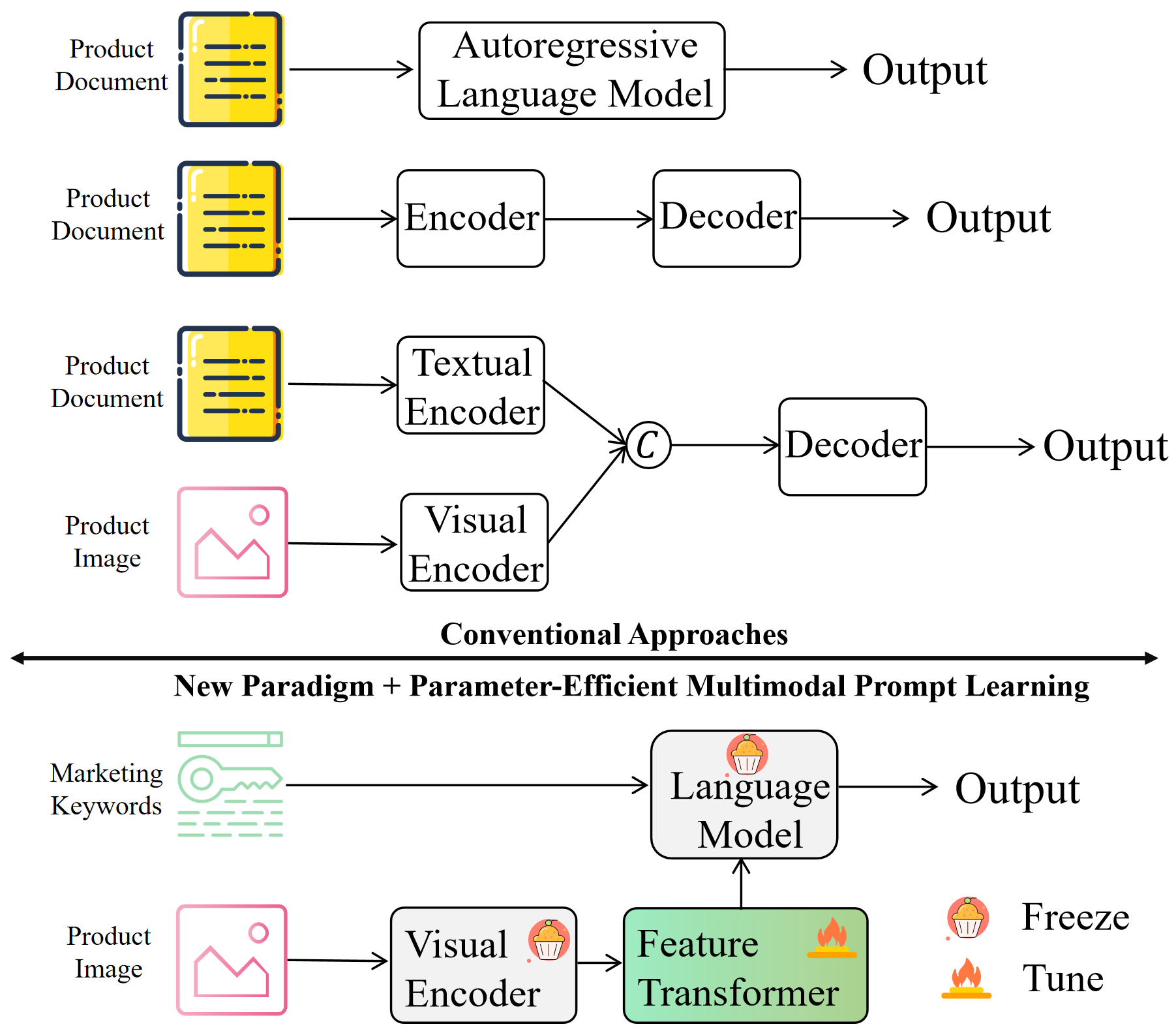}
    \caption{Illustration of conventional approaches and our method for E-commerce product description generation. }
    \label{fig:intro_case}
\end{figure}

With the popularity of online shopping, the E-commerce product description plays a vital role in content marketing and increasing consumer engagement. Automatic generation of product descriptions~\cite{pattern_product, review-based, chan2019stick_prod} has attracted more and more attention, which can be abstracted as a text generation problem from multimodal sources, like Visual Storytelling~\cite{visual_story_tell, huang2016visual}, Image Captioning~\cite{cococaption, Desai_2021_CVPR_caption}, Multimodal Summrization~\cite{zhu2018msmo}, Multimodal Machine Translation~\cite{parida2019hindi}, etc. Previous product description generation works can be divided into two types according to the input source information. One is to generate the corresponding product description from the given long text sequence, as the top two approaches shown in Figure~\ref{fig:intro_case}, which contains product title and its numerous attribute information~\cite{know_pro, Liangke_SymCLKG_TKDE,product_summ,zhan2021probing}, such as color, material, and user's reviews. This type is similar to the task of Abstractive Text Summarization~\cite{Lin_Ng_2019,see-etal-2017-get,parikh-etal-2020-totto,hu2015lcsts}, needing to mine the feature of the product from the long text and generate the corresponding product copy-writing. As the last conventional approach shown in Figure~\ref{fig:intro_case}, the other is to generate the rendering description with product image and its title and various attribute words, which can be abstracted as a text generation problem from multimodal sources, like Multimodal Summarization~\cite{zhu2018msmo}. For the latter, images can provide rich visual information to mine product features, and thus serve as an important basis for product description generation.

In this paper, we suggest generating an E-commerce product description from an image and several marketing keywords. The marketing keywords provide complementary information to the image and contain the product aspects that are difficult to derive directly from the image, such as the product style characteristics, brand, and function. These keywords will explicitly guide the generation of description, i.e., the product description is expected to contain words that are identical to or semantically similar to marketing keywords. 
Compared with the case of inputting a long text sequence containing the product title and various attributes to generate a description, in our task, product marketing features are easy to mine, and the description is naturally controlled by given marketing keywords to some extent. At the same time, keywords are less expensive to be collected by automatic models or humans.

In the realm of product description generation, existing methods~\cite{product_summ, pattern_product, know_pro, deepdepict, peng2022xfboost} employ pretrained image encoders to extract image features, which are combined with keyword encodings and input into a language model for description generation. However, these approaches tend to produce generic and inaccurate descriptions, as they are trained on large-scale datasets, leading to a focus on common words and neglecting product-specific features.
To address this challenge, we introduce an approach called ModICT (Multimodal In-Context Tuning), leveraging the in-context learning and text generation capabilities of language models. Initially, we use a pretrained image encoder to retrieve a similar sample for each input, obtaining representations for both product images. We then employ a learnable feature transformer to convert image features into the language representation space, enabling the incorporation of visual information into the language model. In addition, we input transformed image features, marketing keywords from similar samples, and corresponding descriptions into the language model as in-context references. This way, the pretrained language model learns to generate product descriptions based on similar samples through self-attention mechanisms. During training, we freeze the visual encoder and the generation portion of the language model (e.g., the decoder in a sequence-to-sequence model or all autoregressive model parameters), allowing us to harness the originally powerful generative capabilities of language models for multimodal generation.

To verify the effectiveness of the proposed method, we build a new large-scale E-commerce product description generation dataset, with images and marketing keywords, based on an existing multimodal product summarization corpus~\cite{product_summ}. 
Experimental results show that ModICT outperforms other strong baselines regarding almost all evaluation metrics. Both quantitative and qualitative analyses indicate that ModICT improves the semantic accuracy and diversity of generated descriptions and small language models equipped with ModICT also achieve competitive performances compared to 10x bigger LLMs.

Our contributions are summarized as follows:
\begin{itemize}
    \item We present a product description generation paradigm that is based only on the image and several marketing keywords. For this new setting, we propose a straightforward and effective multimodal in-context tuning approach, named ModICT, integrating the power from the frozen language model and visual encoder.
    \item To the best of our knowledge, our work is the first one to investigate utilizing the in-context learning and text generation capabilities of various frozen language models for multimodal E-commerce product description generation. ModICT can be plugged into various types of language models and the training process is parameter-efficient.
    \item We conduct extensive experiments on our newly built three-category product datasets. The experimental results indicate that the proposed method achieves state-of-the-art performance on a wide range of evaluation metrics. Using the proposed multimodal in-context tuning technical, small models also achieve competitive performance compared to LLMs.
\end{itemize}

\section{Related Work}

\noindent\textbf{Text Generation from Multimodal Sources}.
Text generation tasks from multimodal sources involve the interaction and transformation of information across different modalities. Image captioning~\cite{cococaption,chowdhury2021exploiting,shi2021enhancing, wang2021improving} is a typical image-to-text generation task~\cite{li-etal-2023-neural}, and it requires the model to generate the description of an image. Visual Storytelling~\cite{huang2016visual, visual_story_tell} requires models to generate a long story, given multiple images or a video.
Multimodal Machine Translation~\cite{parida2019hindi, elliott2016multi30k} aims to introduce images to improve the accuracy and diversity of translation, where images could bridge the representation gap across multiple languages. Multimodal Summarization~\cite{li2020multimodal_sentence-summariza, product_summ, jangra2020multi_sum} usually aims at generating a short text to summarize the given long text with relevant images. 
While prior work has mainly focused on generating text from visual information or augmenting text with additional textual content, the exploration of text generation from visual input and keywords remains limited. This unexplored area holds practical potential since short keywords are readily available and can be automatically generated, making them valuable for applications.
\begin{figure*}[t]
    \centering
    \includegraphics[width=0.94\textwidth,height=0.32\textwidth]{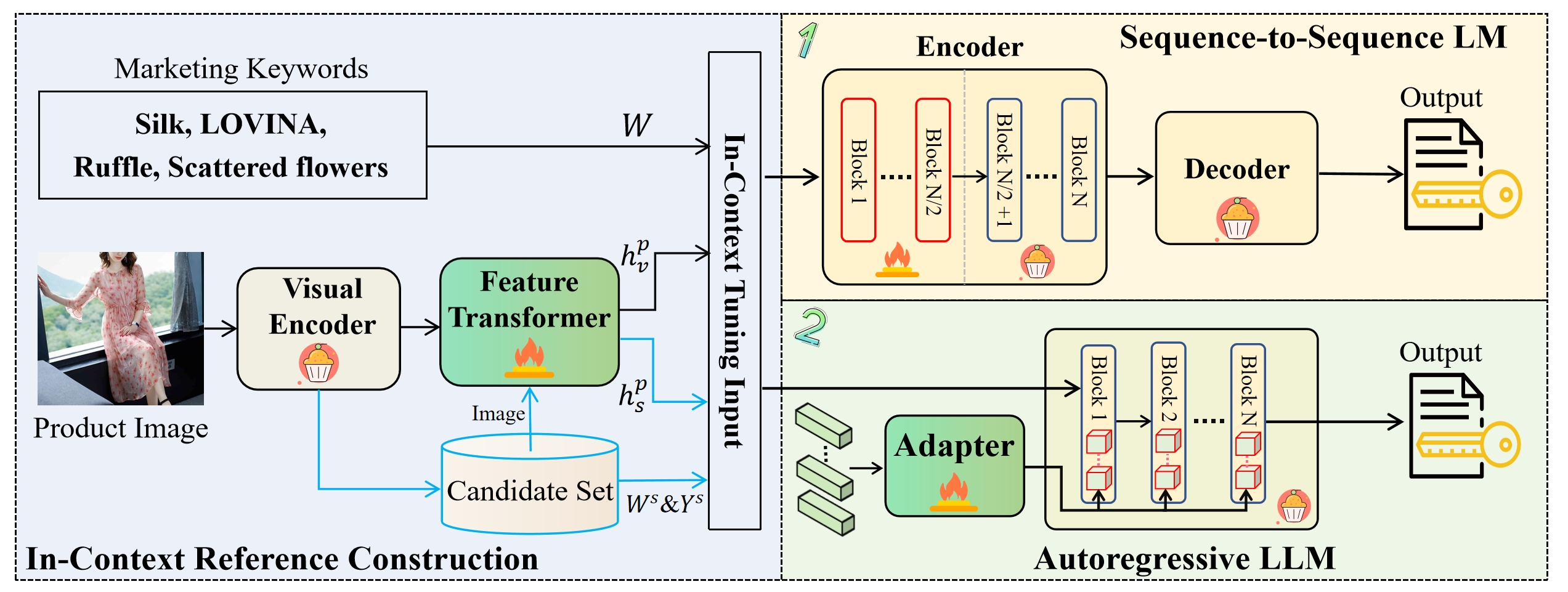}
    \caption{The overall workflow of ModICT. The left part depicts the process of in-context reference construction. The right parts show the efficient multimodal in-context tuning ways for the sequence-to-sequence language model (\textcolor{green}{1}) and autoregressive language model (\textcolor{green}{2}). Blocks with red lines are learnable.}
    \label{fig:model}
\end{figure*}

\textbf{Product Description Generation}. E-commerce product description aims to describe the characteristics of a product in detail and has obtained significant gains in the E-commerce platform. 
To generate detailed, diverse, and accurate descriptions, \citet{pattern_product} propose a pattern-controlled description generation method to control the generation content based on various product properties. They design multiple generation patterns to satisfy different conditions. \citet{know_pro} enhanced product attribute comprehension by incorporating additional knowledge, such as product materials and brand background information. \citet{review-based} utilized customer reviews and clicked product information to diversify generated descriptions.
\citet{xu-etal-2021-k-plug} proposed the K-plug model, a pretrained natural understanding and generation model, demonstrating effective product summarization generation in E-commerce. \citet{zhan2021probing} and \citet{deepdepict} improved description quality using posterior distillation and user preference information. 
In contrast, our approach leverages marketing keywords in conjunction with images, providing complementary guidance for description generation.

\textbf{Vision-assisted Language Models}.
Different from Vision Language Models (VLMs) that are trained with enormous multimodal data such as image-text pairs, vision-assisted language models usually incorporate external visual information into the pretrained language model, which could be used to perform multimodal reasoning or generation. The visual information is obtained by a pretrained visual encoder such as ViT~\cite{vit}, Faster-RCNN~\cite{ren2015faster}, and other variants. Some works~\citep{shi-etal-2019-visually, lu2022imagination, li-etal-2023-neural,li-etal-2023-multi-modal} are proposed to retrieve images from the image corpus and employ visual knowledge to improve the performance of the language model on the downstream tasks. Recently, researchers~\citep{long-etal-2021-generative, yang2021open, zhu2022visualize} utilize the powerful text-to-image technical to obtain the corresponding image of text and inject them into the language model via the prefix-tuning~\citep{li-liang-2021-prefix} way. \citet{li2023blip} and \citet{koh2023grounding, li2023towards,liang2023structure,chen2023temporal,chen-etal-2023-multi} also enable frozen large language models to perform question-answering and image-text retrieval tasks.  This work will explore utilizing frozen large language models to handle the multimodal generation problem in the E-commerce field.
%In a summary, the language model has a significant potential to perform multimodal understanding and generation.

\section{Methodology}

\subsection{Overview}
For the new problem of E-commerce product description generation, the input contains a product image $I$ and corresponding text sequence $W = (w_1, ..., w_i, ..., w_N)$, where $w_i$ represents the $i$-th token of input keyword sequence and $N$ indicates the total number of tokens. The output is the generated description of the product, and we define the ground-truth description as $Y = (y_1, ..., y_i, ..., y_m)$, where $y_i$ and $m$ refer to the $i$-th token and the length of the description, respectively. The proposed ModICT is a parameter-efficient multimodal in-context tuning approach for employing the frozen language models to perform multimodal generation. First, we utilize a frozen visual encoder to retrieve a similar product, which will be used to construct the in-context reference (Sec.~\ref{incontext}). Then, for different types of language models, we adopt two ways to improve the efficiency of multimodal in-context tuning according to the structures of LLMs, which will be shown in Sec.~\ref{learning}. Finally, we briefly present the training strategy of ModICT in Sec.~\ref{training}. 

\subsection{In-Context Reference Construction}
\label{incontext}

We choose samples with visual features similar to in-context references to enhance description diversity. Human-written product descriptions exhibit significant variations, particularly for similar products in the same category. Thus, the language model should imitate human-written text styles and generate diverse, accurate descriptions based on similar products.

\textbf{Selection}. We employ the frozen visual encoder of CLIP~\cite{chinese-clip} to obtain its image representation, denoted as 
$h_{I} = (h_{Ig}, h_{p1}, ...,h_{pn})$, where $h_{Ig}$ and $h_{pi}$ refer to the global and $i$-patch representations of the image.  We consider the same-category training set as the retrieval candidate pool and utilize the same visual encoder to obtain image representations for all products as shown in Figure~\ref{fig:model}. By calculating cosine similarity scores across global representations, we retrieve the most similar product from the same category candidate set. This provides us with the image, marketing keywords, and human-written description of similar products for constructing the one-shot multimodal in-context reference.

\textbf{Construction}.
The obtained image encodings lie in a representation 
space different from the language model due to the
discrepancy between the frozen language model and the image encoder.
To address this issue, we employ a learnable Feature Transformer to convert the image feature into the corresponding language space as the visual prefix vector. Specifically, we feed the global image representation $h_{Ig}$ into a two-layer perceptron with Tanh activation function to obtain the fixed-length visual prefix, denoted as $h_{v}^{p} = (h_{v_1}^{p}, ..., h_{v_L}^{p})$. $L$ is the total length of visual prefix and $h_{v_L}^{p}$ represents the $L$-th prefix embedding.
After obtaining the visual prefixes of two images, we construct a simple in-context template: \textit{``Input Image: <$img$> and Marketing Keywords: $W^s$, output description is $Y^{s}$ \textcolor{blue}{$\backslash$n} Input Image:  <$img$> and Marketing Keywords: $W$, output description is ''}, where <$img$> is a new token to represent the position of visual prefix input and will be frozen during training. To represent the in-context reference, here, we introduce $W^s$, $Y^{s}$, and $h_{s}^p$ to represent the marketing keywords, description, and transformed image feature of the similar sample, respectively. All text inputs (including marketing keywords and the human-written description) are projected into corresponding word vectors via looking up the embedding table of the language model and the representations of <$img$> are added by visual prefix vectors.  The whole sequence representation is input to the following blocks of the language model for multimodal generation.

\subsection{Efficient Multimodal In-Context Tuning}
\label{learning}

Instead of optimizing the overall parameter of LLMs, 
we adopt two parameter-efficient in-context tuning methods according to the structure of LLMs. For sequence-to-sequence language models such as BART and T5, we freeze the decoder and only optimize some parameters of the encoder. In this way, we do not corrupt the generative structure of the language model and do not introduce more parameters.
As the top right shown in Figure~\ref{fig:model}, we optimize the first $N/2$ blocks in the encoder to allow the model to adapt to the multimodal input, where $N$ refers to the total number of blocks in the encoder. For the decoder-only LLMs such as BLOOM~\cite{scao2022bloom}, GPT~\cite{brown2020language_gpt3}, and GLM~\cite{du-etal-2022-glm}, 
inspired by the deep prompt tuning approach \cite{liu2021pv2,tang2022dptdr,wu2022efficient}, we introduce a learnable adapter to allow LLMs to quickly adapt to this multimodal generation task without finetuning any pretrained parameters. Specifically, we randomly initialize $M$ learnable vectors $h^{v} = (h_{1}^{v}, ..., h_{M}^{v})$ and utilize a two-layer perceptron with the ReLU activation function as the adapter to project them into continuous prompts.
The specific calculation process is as follows:
\begin{equation}
    \begin{array}{c}
        \mathbf{h}_{cp} = \mathbf{W}^{a}(ReLU(\mathbf{W}^{1}h^{v} + \mathbf{b}^{1})) + \mathbf{b}^{a}, \vspace{1.0ex}\\
    \end{array}
\label{eq2}
\end{equation}
where $\mathbf{W}^{1}$, $\mathbf{W}^{a}$, $\mathbf{b}^{1}$ and $\mathbf{b}^{a}$ are learnable parameters. The obtained dynamic prompt sequence is $h_{cp} = (h^{1}_{cp}, ..., h^{M*N}_{cp})$, where $N$ is the number of layers of the large language model. As the bottom part shown in Figure~\ref{fig:model},
these dynamic prompts are inserted into each layer of the large model and participate in the self-attention calculation process. The length of inserted vectors for each layer is equal to $M = (M*N)/N$. These continuous prompts are concatenated directly in front of the sequence of input hidden states for each layer of LLMs. Hence, we do not modify any structure of large language models and the training process is parameter-efficient.

\subsection{Training and Inference}
\label{training}

\textbf{Training}. For all models, we adopt the cross-entropy generation loss to train them and the specific process is given in Eq.~\ref{eq3},
\begin{equation}
\begin{aligned}
        \mathcal{L} = -\sum_{i=1}^{m}\mathbf{log}&{P}_{i}(\hat{y}_{i} = y_i | W^{s}, h^{p}_{s}, Y^{s}; \\
        &W, h^{p}_{v}; y_1, ...,y_{i-1}).
\end{aligned}
\label{eq3}
\end{equation}
For autoregressive language models with fewer parameters ($<$1B), we also update their overall parameters due to that continuous prompt tuning is used for efficiently training LLMs. 

\noindent\textbf{Inference}. For each testing sample, the corresponding similar sample is retrieved from the training set. The inferring process is similar to Eq.~\ref{eq3} and is equipped with some common generation methods, e.g., beam sample strategy.

\section{Experiment}

\subsection{Dataset: MD2T}

\textbf{Construction}. In previous product description generation tasks, input data often included product titles, attributes, images, and many others. Some information was redundantly reflected in both text and images. To address this, we build a new product description generation dataset with images and keywords, named MD2T~\footnote{We will release the preprocessing codes and data sources.}, based on the large-scale millions of multimodal Chinese E-commerce product summarization corpus~\cite{li2020aspect_summ}. 
To be specific, we collect the product style, brand, color, material, and popular element dictionaries from the released text-based E-commerce product summarization dataset~\cite{yuan-etal-2020-faithfulness}. The collected product attribute aspects of Cases \& Bags. We then use the Chinese word segmentation tool Jieba\footnote{https://github.com/fxsjy/jieba} with the above dictionary to segment the long text sequence of product samples. For the obtained word segmentation set, we filter out words that can be easily derived from images (e.g., color, size, and shape) via word matching. We select the style, popular element, brand, material, and a few other randomly sampling words ($20\%$ of the number of other remaining words) from the remaining word set as the marketing keywords. Finally, we remove instances whose product descriptions do not contain any marketing keywords (exact matching) to ensure that marketing keywords can guide the description generation. 

\begin{table}[t]
\renewcommand\arraystretch{1.0}
\setlength\tabcolsep{1.2pt}
\footnotesize
\centering

\begin{tabular}{lcccc}
\hline
\multicolumn{1}{l}{\textbf{MD2T}} & 
\multicolumn{1}{c}{\textbf{Cases\&Bags}} &
\multicolumn{1}{c}{\textbf{Clothing}} & 
\multicolumn{1}{c}{\textbf{Home Appliances}} \\ \hline
\#Train & $18,711$ & $200,000$ & $86,858$ \\
\#Dev & $983$ & $6,120$ & $1,794$ & \\
\#Test & $1,000$ & $8,700$ & $2,200$ & \\ 
Avg$_N$\#MP & $5.41$ & $6.57$ & $5.48$ \\
Avg$_L$\#MP & $13.50$ & $20.34$ & $18.30$  \\
Av$_L$\#Desp & $80.05$ & $79.03$ & $80.13$\\ 
\hline
\end{tabular}
\caption{The detailed statistics of MD2T. Avg$_N$ and Avg$_L$ represent the average number and length respectively. MP and Desp indicate the marketing keywords and description.}
\label{dataset}
\end{table}

\textbf{Statistic}. The detailed statistics are shown in Table~\ref{dataset}. 
The total number of samples across the three categories is approximately 300,000. Product descriptions are relatively long, containing about five marketing keywords and spanning around 80 words. The keyword length varies across categories, with Clothing and Home Appliances having longer keywords compared to Bags \& Cases. This difference arises from challenges in collecting category-specific dictionaries for keyword segmentation and filtering, such as brand, style, and material. While some samples may contain a few noisy words (up to 2), they do not significantly impact our overall analysis.

%As we can see, the total number of samples for the three categories is about $300,000$ and the rendering description of products is a long text with several marketing keywords (about five keywords), about $80$ words in length. The length of keywords in the categories of Clothing and Home Appliances is longer than that in the category of bags \& cases. This is mainly attributed to the fact that it is difficult to collect category-related dictionaries for keyword segment and filtering, such as brand, style, and material. Although these samples contain a few noisy words (no more than $2$), it does not affect the overall experimental analysis.

\begin{table*}[t]
    \renewcommand\arraystretch{1.0}
    \setlength\tabcolsep{1.0pt}
    \footnotesize
    \centering

    \begin{tabular}{ccc|ccccc|cccc}
    \hline
    Model & \#TunedPara & \#TotalPara & B@1$\uparrow$ & B@2$\uparrow$ & R@1$\uparrow$ & R@L$\uparrow$ & BS$\uparrow$ & D-2 $\uparrow$& D-3 $\uparrow$& D-4 $\uparrow$& D-5 $\uparrow$
    \\
    \hline

    MMPG+D~\cite{product_summ} &96M&96M & 30.91 & 10.0 & 30.92 & 10.98 & 32.0 &-&-&-&-\\
    
    MMPG+D+C~\cite{product_summ}&100M&100M & 33.31& 10.72& 31.47& 21.25 & 31.8 &2.66	&4.73	&7.55&	10.54\\
    
    M-kplug~\cite{xu-etal-2021-k-plug} & 225M &225M & 30.96& 8.76& 29.43 & 19.45& 30.0 & 9.34&16.54&22.57 &27.48\\
    
    Oscar~\cite{zhang2021mengzi} & 110M &110M & 29.52& 7.58& 28.32& 17.93 &
    28.2 &-&-&-&-\\
    
    Oscar-GPT~\cite{pmlr-v139-cho21a} &279M&279M& 33.70 &	10.81 & 32.12 &	21.19 & 32.1 &11.35	&22.68	&33.21	&44.23\\
    \hline
    ModICT (BART-L) &232M & 521M & \textbf{36.54}	& \textbf{13.28}	&\textbf{35.43}&\textbf{24.50} & 34.9 & 12.76 & 24.11 & 36.23 & \textbf{47.31}\\
    w/o MICT &232M & 521M & \text{36.30}	& \text{13.15}	&	\text{35.39}	&	\text{24.45}	&34.7 & 12.35 & 23.38 & 34.51 & 45.21\\ 
    w/o MICT (full) &435M & 521M & 34.95 & 11.94 & 33.63 & 22.93 & 33.4 & 10.59 & 20.34 & 31.58 & 42.34\\ \hline
    %ModICT (BART-L)$^{*}$ &232M & 521M & +2.84 & +2.47 & +3.21 & +3.31 & +2.8 & +1.41&1.43 & +2.01 & +3.08\\
    %-MICT &232M & 521M & \text{36.30}	& \text{13.15}	&	\text{35.39}	&	\text{24.45}	&34.7 & 12.35 & 23.38 & 34.51 & 45.21\\ 
    %-MICT(full) &435M & 521M & 34.95 & 11.94 & 33.63 & 22.93 & 33.4 & 10.59 & 20.34 & 31.58 & 42.34\\
    ModICT (BART-RD) & 181M & 874M & 35.40 & 12.56 & 34.10& 23.92 & 33.9& 13.07 & 23.86 & 34.40 & 43.67\\
    %-MICT & 181M & 874M & 36.01 & 12.74 & 34.40 &  24.16 & 34.0 & 9.25 & 17.60 & 27.42 & 37.00\\ \hline 
    ModICT (GLM-L) &364M & 450M & 28.20 & 10.28 & 32.43& 22.55 & 32.1& 10.86 & 20.85 & 30.17 & 38.67\\
    ModICT (GLM-10B)  &511M & 10.6B &28.83 & 10.35 & 32.15 & 22.78 & 32.7 & 12.20 & 23.06 & 32.56 & 41.11\\
    %\hline
    ModICT (BLOOM-1.1B) &118M &1.4B & 32.61 & 11.88 & 34.08 & 24.23 & 34.6 & 13.23 & 24.07 & 35.38 & 45.74\\
    %-Adapter & 42M & 1.4B & 31.10 & 11.06 & 32.90 & 24.30 & 32.0 & 13.47 & 24.54 & 35.25 & 44.64\\
    %-Adapter-MICT & 42M & 1.4B & 30.96 & 11.16 & 33.16 & 23.90 & 31.5 & 8.58 & 16.00 & 24.74 & 33.12\\\hline
    ModICT (BLOOM-1.7B) &156M & 2B & 33.62 & 12.07 & 34.50 & 24.31 & 34.5 & 11.12 & 19.89 & 29.14 & 37.83\\
    %-Adapter & 55M & 2B & 32.36 & 11.62 & 33.70 & 24.28 & 33.5 & \text{13.73} & \text{25.07} & 36.17 & 46.08\\
    %-Adapter-MICT & 55M & 2B & 31.79 & 11.18 & 33.38 & 23.92 & 31.6 & 8.96 & 16.82 & 25.96 & 34.75 \\\hline
    ModICT (BLOOM-3B) &226M & 3.4B & 33.15 & 12.08 & 33.88 & 24.06 & 34.4 & 12.91 & 23.68 & 34.60 & 44.62\\
    %-Adapter & 69M & 3.4B & \\
    %-Adapter-MICT & 69M & 3.4B & \\\hline
    ModICT (BLOOM-7.1B) & 360M & 7.6B& 32.86 & 12.06 & 34.48 & 24.28 & $\mathop{\textbf{35.3}}$ & 12.95 & 23.42 & 34.33 & 44.46\\
    w/o Adapter & 108M & 7.6B & 32.36 & 12.18 & 34.28 & 24.10 & 34.4 & \textbf{13.8} & \textbf{25.08} & \textbf{36.45} & \text{46.85}\\
    w/o Adapter+MICT & 108M & 7.6B & 32.00 & 11.62 & 33.84 & 24.19 & 33.1 & 10.0 & 18.45 & 28.0 & 37.26 \\
    %-Adapter & 108M & 7.6B & 32.36 & 12.18 & 34.28 & 24.10 & 34.4 & \textbf{13.8} & \textbf{25.08} & \textbf{36.45} & \text{46.85}\\
    %-Adapter-MICT & 108M & 7.6B & 32.00 & 11.62 & 33.84 & 24.19 & 33.1 & 10.0 & 18.45 & 28.0 & 37.26 \\
    %\limits _{\textbf{+3.2}}
    \hline
    \end{tabular}
        \caption{Automatic evaluation on the testing set of Cases\&Bags. The bold indicates the best performance. ``\#TunedPara'' and ``\#TotalPara'' represent the trainable and overall parameters of models respectively. ``BS'' refers to the BertScore evaluation metric. ``MICT'' represents the proposed multimodal in-context tuning way. ``full'' indicates that the overall parameters of language models are tuned during training. ``Adapter'' shows the parameter-efficient prefix tuning method for autoregressive LLMs. } 
    \label{tab:results_bags}
\end{table*}

\begin{table*}[t]
    \renewcommand\arraystretch{1.0}
    \setlength\tabcolsep{1.2pt}
    \footnotesize
    \centering
    \begin{tabular}{ccc|ccccc|cccc}
    \hline
    Model & \#TunedPara & \#TotalPara & B@1 & B@2 & R@1 & R@L& BS & D-2 & D-3 & D-4 & D-5\\
    \hline
    
    MMPG+D~\cite{product_summ}&96M&96M  & 24.94	& 6.22 &26.51 &	17.49 &22.1 &- &- &- &-\\
    
    MMPG+D+C~\cite{product_summ} &100M&100M  & 
    25.04& 7.05 &26.12	&18.16 &24.8 &- &- &- &- \\
    
    M-kplug~\cite{xu-etal-2021-k-plug} & 225M & 225M
    & 31.64	& 10.35	&30.48	&20.42	& 29.7 & 10.03 & 22.00 & 34.60 & 45.81\\
    
    Oscar~\cite{zhang2021mengzi} & 110M &110M 
    & 27.91 &	6.89	&	26.61	&	16.51 &	25.2 &- &- &- &-\\
    
    Oscar-GPT~\cite{pmlr-v139-cho21a} &279M&279M
    &31.22 &	10.15	&	30.22	&	20.27 &	29.6 & 12.22	&24.15	&36.43 & 47.41\\\hline
    %& 34.93& 12.30 & 33.34& 21.91&	34.6 
    ModICT (BART-L) &232M & 521M & 
    \textbf{34.27}	& \textbf{12.56}	&	\textbf{33.24}	&\textbf{23.42} & \textbf{32.1} & 14.73 & 29.36 & 44.24 & 56.82 \\
    w/o MICT &232M & 521M &  32.57 & 11.43 & 32.04 & 22.90 & 30.7 &9.34 & 18.82 & 30.43 & 41.28\\
    w/o MICT (full) &435M & 521M & 33.56 & 11.89 & 32.73 & 23.08 & 31.5 & 9.01 & 17.11 & 29.17 & 39.98\\  \hline
    ModICT (BART-RD) & 181M & 874M & 30.62 & 10.47 & 31.75 & 22.05 & 29.7& 16.20 & 32.27 & 46.73 & 58.25\\
    %w/o MICT & 181M & 874M & 30.16 & 10.10 & 31.16 & 21.92 & 28.8 & 9.28 & 18.92 & 30.37 & 42.19\\ \hline
    %\hline
    ModICT (GLM-L) & 364M & 450M & 26.07 & 9.65 & 29.40 & 20.43 & 29.1 & 12.62 & 25.74 & 37.46 & 47.41\\
    ModICT (GLM-10B)  & 511M & 10.6B & 25.75 & 9.11 & 29.72 & 20.88 & 29.5 & 13.23 & 27.31 & 39.73 & 50.11 \\ 
    %\hline
    ModICT (BLOOM-1.1B) &118M &1.4B & 30.10 & 10.45 & 30.99 & 22.25 & 29.9 & 13.97 & 27.80  &41.81  &53.71\\
    %-Adapter  &42M &1.4B & 29.48 & 10.18 & 30.36 & 21.90 & 28.9 & 17.30 & 34.10 & 48.26 & 59.26\\
    %-Adapter-MICT  &42M &1.4B & 24.52 & 8.07 & 28.79 & 21.60 & 24.7 & 9.54 & 19.10 & 29.38 & 38.46\\
    %\hline
    ModICT (BLOOM-1.7B) &156M & 2.0B & 30.25 & 10.65 & 31.21 & 22.24 & 30.3 & 15.16  & 29.57  &43.30  &54.64 \\
    %-Adapter  & 55M & 2.0B & 29.02 & 10.00 & 30.08 & 21.77 & 28.4 & 17.46 & 34.14 & 48.29 & 59.26\\
    %-Adapter-MICT  & 55M & 2.0B & 26.74 & 8.91 & 29.44 & 22.02 & 26.0 & 10.49 & 20.69 & 31.24 & 40.65 \\

    ModICT (BLOOM-3B) &226M & 3.4B & 29.77 & 10.54 & 31.14 & 22.47 & 30.2 & 14.46 & 28.31  &42.11  &53.81\\
    ModICT (BLOOM-7.1B) & 360M & 7.6B  & 30.91 & 10.89 & 31.46 & 22.20 & 30.5 & 14.12  &28.09  &42.12  &53.88\\
    -Adapter  & 108M & 7.6B & 30.36 & 10.80 & 31.42 & 22.50 & 30.3 & \textbf{17.63} & \textbf{34.31} & \textbf{48.89} & \textbf{60.50}\\
    -Adapter-MICT & 108M & 7.6B & 28.68 & 9.64 & 29.63 & 21.99 & 27.1 & 10.72 & 21.10 & 31.99 & 41.69\\
    \hline
    \end{tabular}
    \caption{Automatic evaluation on the testing set of Home Appliances.}
    \label{tab:result_home_app}
\end{table*}

\subsection{Experimental Settings}
\textbf{Evaluation Metrics}. In our experiments, we adopt the widely-used automatic overlap-based metrics BLEU~\cite{papineni2002bleu} and ROUGE~\cite{lin-2004-rouge} in the text generation area to evaluate the generated description. However, these word-overlap based metrics ignore the contextual semantic discrepancy. Therefore, we use a popular word embedding-based similarity evaluation metric BERTScore~\cite{zhang2019bertscore}, dubbed as BS, to capture the fine-grained semantic difference, which employs the contextualized representation of words. We adopt the pretrained BERT-base-Chinese parameters to initialize the model for BERTScore. We also design a simple yet effective evaluation metric to measure the whole \textbf{diversity} of models in the testing set. Specifically, we concatenate all generated descriptions into a long sequence and remove punctuation. Then this sequence is converted into a list $L$ based on Chinese character segmentation. The whole list sequence is divided by the Distinct N-gram segmentation method~\cite{li2015diversity} and thus we can obtain the no repeated n-gram word set $S_n$, where the maximum number of $S_n$ does not exceed the length of $L$.
The n-gram word diversity of generated descriptions can be calculated as follows: $D\text{-}n = \frac{Number\_of(S_n)}{Length\_of(L)} *100$.
In this way, $D\text{-}n$ can intuitively show the diverse generation capability of a model.

\begin{table*}[t]
    \renewcommand\arraystretch{1.0}
    \setlength\tabcolsep{1.2pt}
    \centering
    \footnotesize

    \begin{tabular}{ccc|ccccc|cccc}
    \hline
    Model & \#TunedPara & \#TotalPara & B@1 & B@2 & R@1 & R@L& BS & D-2 & D-3 & D-4 & D-5\\
    \hline
    MMPG+D~\cite{product_summ}&96M&96M  & 29.01 & 8.30 & 27.03 & 17.92 & 30.2 & -& -&-& -\\
    
    MMPG+D+C~\cite{product_summ} &100M&100M & 29.63 & 8.65 & 27.61& 18.26 & 30.5 & 1.63	&2.24	&4.01	&5.28\\
    
    M-kplug~\cite{xu-etal-2021-k-plug} &225M & 225M & 33.38 & 10.91 & 31.42& 20.34& 33.0 & 4.5 & 10.05 &  18.54 & 27.22 \\
    
    Oscar~\cite{zhang2021mengzi} & 110M &110M & 29.69 & 7.74& 28.38 & 16.71 & 29.4 & - & - & -& - \\
    
    Oscar-GPT~\cite{pmlr-v139-cho21a} &279M&279M & 33.54	& 10.98  &	31.92 & 20.51	& 33.1 & 5.1 & 12.18 & 20.31 & 28.15\\

\hline
    %& 34.93& 12.30 & 33.34& 21.91&	34.6 
    ModICT (BART-L) &232M & 521M & \textbf{35.07} & \textbf{12.57} & \textbf{33.71} & \textbf{22.52} & \textbf{34.9}&
    4.17 & 9.28 & 18.05 & 24.45\\
    w/o MICT &232M & 521M & \text{34.93} & \text{12.30} & \text{33.34} & \text{21.91}& 34.6  & 2.95 &6.84 & 12.92 & 20.0 \\
    ModICT (BART-RD) & 181M & 874M & 33.48& 11.17& 32.41 & 21.86 & 33.3 & 5.82 &\text{13.45} &22.48 &31.54\\
    %w/o MICT & 181M & 874M & 33.25 & 11.28 & 32.16 & 21.46 & 33.0 & 2.68 & 6.08 &11.39& 17.71\\ 
    %\hline
    ModICT (GLM-L) & 364M & 450M & 26.03 & 9.10 & 29.51 & 19.44 & 31.4 & 3.67 &8.71 &14.57 &20.62 \\
    ModICT (GLM-10B)  & 511M & 10.6B & 26.45& 9.21 & 29.66 & 19.74 & 31.8 & \textbf{6.71} & \textbf{17.18} & \textbf{28.16} & \textbf{37.61}\\ 
    ModICT (BLOOM-1.1B) &118M &1.4B & 29.14 & 10.07 & 31.59 & 21.40 & 34.2 & 4.07 & 9.27 & 16.84 & 25.04\\
    ModICT (BLOOM-1.7B) &156M & 2.0B & 30.25 & 10.17 & 31.26 & 20.83 & 33.6 & 3.75 & 8.16 & 14.25 & 20.76\\
    ModICT (BLOOM-3B) &226M & 3.4B & 29.34 & 10.19 & 31.56 & 21.15 & 33.9 & 4.39 &9.58 &17.18 &25.17 \\
    \hline
    ModICT (BLOOM-7.1B) & 360M & 7.6B & 30.58 & 10.52 & 32.12 & 21.65 & \textbf{34.9} & 4.64 & 9.13 & 16.44 & 25.21\\
    w/o Adapter &  108M & 7.6B & 30.45 & 10.49 & 31.91 & 21.36 & 34.2 & \text{5.97} & 13.21 & \text{22.71} & \text{32.08} \\
    w/o Adapter+MICT & 108M & 7.6B & 30.16 & 10.50 & 31.61 & 21.05 & 33.1 & 3.39 & 7.67 & 14.43 & 21.89\\
    \hline
    \end{tabular}
        \caption{Model performances (accuracy and diversity) on the testing set of Clothing.}
    \label{tab:result_clothing}
\end{table*}

%A robust and high-performance model should describe different products in a variety of languages to fulfill the fact that different products have various characteristics and visual impacts. It is also important that diverse descriptions are applicable in practice. 

\textbf{Comparative Models}.
We compare our method with several SOTA multimodal product description generation methods.
MMPG \cite{product_summ} is a multimodal E-commerce product summarization model based on LSTM~\cite{hochreiter1997long}. We use it with the DecInit~(D) and \text{Copying~(C)} mechanism on this task, and \citet{product_summ} have verified their effectiveness to improve the quality of generated results and \text{MMPG + D + C} performs the best on the Case \& Bag category in multimodal E-commerce product summarization~\cite{li2020aspect_summ}.
\text{M-kplug} is an extension of the text-based pretrained model k-plug~\cite{xu-etal-2021-k-plug} in E-commerce, which injects the visual signals into the decoder layer.
\text{Oscar}~\cite{li2020oscar, zhang2021mengzi} is a transformer-based pretrained conditional cross-modal model, having achieved great success in many vision-language tasks. 
%We adopt the pretrained parameters released by \citet{zhang2021mengzi}, which have been trained on a large-scale Chinese image-language dataset.
\text{Oscar-GPT}~\cite{kayser2021vil} is a sequence-to-sequence vision-language generation model. 
%Its encoder adopts Oscar to encode the cross-modal information, and the decoder generates the text with attention to the encoding sequence.

\textbf{Implementation Details}.
We train all models on two A100-40G GPUs with the python environment. We train models with an initial learning rate $1e^{-4}$ and the learning rate declines via the linear way. The total training steps are $10$ epochs with $1,000$ warm-up steps. For baselines and ModICT variants equipped with very small autoregressive language models (like GLM-L\cite{du-etal-2022-glm}), we update the overall parameters of language models. The batch sizes of training and inference are set to $32$ and $10$ respectively. We adopt the Chinese CLIP-ViT-16~\cite{chinese-clip} as the image feature extractor, which contains 84M parameters.
We use the validation set to select the best parameter when training all models. For inference, we adopt the beam sample generation method and set the beam and sample sizes to $4$ and $20$, respectively. The visual prefix length $L$ is set to 5 and the length of continuous prompts ($M$) is set to 10. GLM-10B and BLOOM-7B are trained in mixed precision (half-precision for the forward/backward computations, full-precision for the gradient update) with the AdamW optimizer. We test various models with different parameter sizes on our newly built dataset of three categories of products.
%The image size is adjusted to $224\times224$ and each patch is set to $16\times16$, so the length of the patch sequence is $196$.
%In this paper, we mainly perform experiments on two autoregressive language models (multilingual BLOOM~\cite{scao2022bloom} and Chinese GLM~\cite{du-etal-2022-glm}) and Chinese encoder-decoder language model BART~\cite{km-bart} and BART-RD~\cite{fengshenbang}. 

\subsection{Quantitative Analysis}

%``*'' indicates that the performance of corresponding ModICT version compares to the best baseline in terms of evaluation metrics.

%CLIP-V 86M Visual Mapping Network 29M  total 115M
%\multicolumn{3}{c|}{} &\multicolumn{5}{c|}{Clothing} &\multicolumn{5}{c}{Home Appliances} \\

\textbf{Content Accuracy}. 
Evaluation results in Tables~\ref{tab:results_bags}, \ref{tab:result_home_app}, and \ref{tab:result_clothing} reveal ModICT(BART-L) excels in content accuracy across most metrics, significantly outperforming Oscar-GPT and M-kplug (e.g., R@L $\uparrow$ 3.31, BERTScore $\uparrow$ 2.8). Frozen autoregressive BLOOM also outperforms baselines in essential metrics (R@L, BERTScore). This demonstrates the effectiveness of the learnable multimodal in-context tuning approach across various language models. Additionally, BLOOM outperforms GLM-10B in all product categories, despite its larger parameter count, possibly due to its multilingual training data. Increasing language model parameters slightly enhances performance in these fixed E-commerce product descriptions, achieved through the proposed in-context tuning approach. As the parameters of the language model increase, the performance of models on the three product categories increases slightly, which may be due to the fixed description paradigm of E-commerce products, and the small model could acquire corresponding generation capability through the proposed multimodal in-context tuning approach.

%We present evaluation results of all models in Tables~\ref{tab:results_bags}, \ref{tab:result_home_app}, and \ref{tab:result_clothing}. By comparing all model variants with baselines, it is observed that the ModICT(BART-L) performs best regarding most evaluation metrics of content accuracy and significantly outperforms performing-best baseline Oscar-GPT such as R@L: $\uparrow$ 3.31, $\uparrow$ 3.15, and $\uparrow$ 2.01 on three product categories; BERTScore: $\uparrow$ 2.8, $\uparrow$ 2.5, and $\uparrow$ 1.8. Additionally, we observe that frozen autoregressive BLOOM also achieves superior performance on important evaluation metrics (R@L and BERTScore) compared to baselines. Hence, the simple multimodal in-context learning approach is capable of generalizing to various language models. Compared with GLM (10B), BLOOM performs better in all product categories, which may be attributed to the self-capability of LLMs. Although GLM-10B has a larger number of parameters, it is only trained on the Chinese corpus, yet BLOOM is trained on a larger multilingual corpus. As the parameters of the language model increase, the performance of models on the three product categories increases slightly, which may be due to the fixed description paradigm of E-commerce products, and the small model could acquire corresponding generation capability through the proposed multimodal in-context learning approach. 

\begin{table}[t]
\renewcommand\arraystretch{1.0}
\setlength\tabcolsep{1.4pt}
\centering
\footnotesize

\begin{tabular}{c|ccc|ccc}
    \hline
     \multicolumn{1}{c}{\textbf{TrainS}}
     & \multicolumn{1}{|c}{\textbf{R@1}} 
     & \multicolumn{1}{c}{\textbf{R@L}} 
     & \multicolumn{1}{c}{\textbf{BS}} 
     & \multicolumn{1}{|c}{\textbf{D-3}} 
     & \multicolumn{1}{c}{\textbf{D-4}} 
     & \multicolumn{1}{c}{\textbf{D-5}} \\ 
    \hline
    200k & \textbf{33.71} & \textbf{22.52} & 34.9 & 9.28 & 18.05 & 24.45\\
    50k & 33.30 & 22.10 & 34.6 & \textbf{10.64} & \textbf{18.65} & 27.13\\
    40k & 33.54 & 22.16 & \textbf{34.9} & 10.37 & 18.57 & \textbf{27.57}\\
    30k & 33.49 & 22.11 & 34.8 & 9.86 & 17.55 & 25.89\\
    \hline
\end{tabular}
\caption{ModICT(BART-L) performances on Clothing with various scales of training samples . ``TrainS'' refers to the size scale of training samples.}
\label{tab:result_small}
\end{table}

%Diversity evaluation aims to assert the diversity generation capability of the framework to produce descriptions of similar products.
\noindent\textbf{Diversity}.
To assess diversity, we analyze Tables \ref{tab:results_bags}, \ref{tab:result_home_app}, and \ref{tab:result_clothing}. ModICT variants show varying performance across product categories but consistently outperform strong baselines (M-kplug and Oscar-GPT) in diversity evaluation. In Cases \& Bags, ModICT (BART-L) improves D-5 score by 3.08, while ModICT(BLOOM-7B) and ModICT (BART-L) achieve impressive improvements of \textbf{3.93} and \textbf{9.41} for Clothing and Home Appliances. Notably, ModICT variants perform less well in Clothing, likely due to overfitting on common words in large-scale training sets, especially for variants with more parameters. Table \ref{tab:result_small} reveals that ModICT (BART-L) achieves superior content accuracy and diversity in Clothing with fewer training samples, showcasing the feasibility of training small models using multimodal in-context tuning for practical, cost-effective applications.

%After analyzing the evaluation results presented in Tables \ref{tab:results_bags}, \ref{tab:result_home_app}, and \ref{tab:result_clothing}, we can observe that the ModICT variants exhibit different performances across the three product categories. Nevertheless, their diversity evaluation results surpass strong baselines (M-kplug and Oscar-GPT) by a significant margin. For instance, in the Cases\&Bags, ModICT (BART-L) achieves a D-5 score improvement of 3.08, while ModICT(BLOOM-7B) and ModICT (BART-L) obtain D-5 score improvements of 3.93 and \textbf{9.41}, respectively, for Clothing and Home Appliances. It is noteworthy that the ModICT variants perform worse on Clothing than on Home appliances and Cases\&Bags. This can be attributed to the fact that the large-scale training set causes LLMs to overfit to common words, particularly for ModICT variants with more learnable parameters. Table~\ref{tab:result_small} shows that ModICT (BART-L) attains superior content accuracy and diversity by training on fewer data samples in the Clothing category. This suggests that we can train models by using multimodal in-context learning with a relatively small number of samples. Therefore, in practical applications, we can deploy a system with small language models trained using multimodal in-context learning, resulting in minimal cost overhead.

\begin{table}[t]
\renewcommand\arraystretch{1.0}
\setlength\tabcolsep{1.2pt}
\centering
\footnotesize

    \begin{tabular}{lc|cc|ccc}
        \hline
         \multicolumn{1}{c}{\textbf{Model}}
         & \multicolumn{1}{c}{\textbf{NRE}}
         & \multicolumn{1}{|c}{\textbf{R@L}} 
         & \multicolumn{1}{c}{\textbf{BS}} 
         & \multicolumn{1}{|c}{\textbf{D-3}} 
         & \multicolumn{1}{c}{\textbf{D-4}} 
         & \multicolumn{1}{c}{\textbf{D-5}} \\ 
        \hline
        BLOOM-1.7B &1-shot & 22.24 & 30.3 & 29.57 & 43.40 & 54.64\\
        BLOOM-1.7B & 2-shot & 22.07 & 29.3 & 30.75  & 44.40 & 55.60\\
        BLOOM-1.7B & 3-shot & 21.70 & 28.0 & 30.40  & 44.10 & 55.40\\
        BLOOM-1.7B$^{*}$ & 0-shot & 21.50 & 26.0 & 20.68 & 31.24 & 40.64\\
        BLOOM-1.7B$^{*}$ & 1-shot & 19.52 & 26.3  & 41.16 & 55.29 & 64.53\\
        BLOOM-1.7B$^{*}$& 2-shot & 19.73 & 27.3  & 40.23 & 54.39 & 63.96\\
        BLOOM-1.7B$^{*}$ & 3-shot & 19.63 & 26.6  & 38.78 & 52.74 & 62.54\\
        \hline
        BLOOM-7.1B &1-shot & 22.20 & 30.5 & 28.09 & 42.12 & 53.88\\
        BLOOM-7.1B & 2-shot & 21.94 & 30.3 & 32.21  & 45.88 & 56.80\\
        BLOOM-7.1B & 3-shot & 21.83 & 30.0 & 33.65 & 47.48& 58.56\\
        BLOOM-7.1B$^{*}$ & 0-shot & 21.99 & 27.1 & 21.10 & 31.99 & 41.69\\
        BLOOM-7.1B$^{*}$ & 1-shot & 20.15 & 26.8  & 42.20 & 56.38 & 65.85\\
        BLOOM-7.1B$^{*}$& 2-shot & 20.12 & 24.8  & 36.77 & 50.42 & 60.40\\
        BLOOM-7.1B$^{*}$ & 3-shot & 19.80 & 22.1 & 36.76 & 50.41 & 60.52\\
        \hline
        
    \end{tabular}
    \caption{Model performances with various in-context reference examples on the testing set of Home Appliances. $^{*}$ refers to that the corresponding model without adapter are not trained with MICT. ``NRE'' refers to the Number of Reference Examples. }
        \label{tab:result_shots}
\end{table}

\subsection{Ablation Study}

\textbf{Effectiveness of ModICT}. From all experimental Tables, it is observed that, in the case of both the sequence-sequence language model and the autoregressive large language model, the MICT mainly improves the diversity of generated content. For various ModICT (BLOOM) variants, it also advances the content accuracy and substantially promotes the diversity of content (model performance comparison: -Adapter vs. -Adapter-MICT), especially for the category of Home Appliances. 

\noindent\textbf{Impact of Adapter}. By ablation experiments on ModICT(BLOOM), we observe that the adapter mostly improves the overall content accuracy yet sometimes leads to a slight decrease in diversity. It may be attributed to that more parameters are introduced and we add the continuous prompts in each layer of LLMs. The performance comparisons between ModICT (BLOOM) and its ``-Adapter-MICT'' variant indicate that it is useful for improving the content diversity and accuracy by introducing the adapter and MICT together.

\noindent\textbf{Tuned Parameter vs. Performance}. 
Small language models (<1B) also excel in high-quality product description generation through multimodal in-context tuning. However, when fine-tuning generation-related parameters, diversity decreases (e.g., -MICT vs. -MICT(full) in Tables~\ref{tab:results_bags}, ~\ref{tab:result_home_app}, and ~\ref{tab:result_clothing}), and content accuracy declines when training the overall BART-L parameters for Cases \& Bags. To enhance overall accuracy and diversity, we recommend freezing the LLMs and using multimodal in-context tuning approach with one-shot reference.

%According to the performance analyses of content accuracy and diversity, we could know that small language models (< 1B) also achieve the high-quality product description generation capability under the proposed multimodal in-context learning paradigm. Secondly, when we finetune the generation-related parameters, the diversity performance of models will decrease, e.g., -MICT vs. -MICT(full) in Tables and the content accuracy also declines when we train the overall parameter of BART-L on the training set of cases \& Bags. To improve the overall accuracy and diversity, we suggest freezing the LLMs and employing the multimodal in-context learning method.

\noindent\textbf{Training Data Size}. In Table~\ref{tab:result_small}, ModICT (BART-L) trained on 40k samples achieves content accuracy similar to the 200k-sample model but with better diversity. Comparing all models on the 18k sets of Clothing and Cases \& Bags, they perform better in Cases \& Bags with a small-scale training set. This suggests that our multimodal in-context tuning approach is effective with limited labeled data.

\noindent\textbf{Analysis of In-Context References}. 
Table~\ref{tab:result_shots} displays LLM performance with varying in-context examples. ModICT improves content accuracy and maintains diversity compared to ModICT(BLOOM)$^{*}$ with one-shot input. As in-context samples increase, ModICT diversity rises while content accuracy slightly decreases, in line with our motivation to enhance description diversity. However, it highlights the instability of large multi-modal models based on LLMs. Increasing LLM parameters results in a wider range of outcomes, yielding high diversity but lower accuracy. It may be attributed to the fact that the generation of larger language models is more diverse and less controllable.

\begin{table}[t]
\renewcommand\arraystretch{1.0}
\setlength\tabcolsep{1.5pt}
\centering
\footnotesize

    \begin{tabular}{lcccc}
        \hline
         \multicolumn{1}{c}{\textbf{Model}} & \multicolumn{1}{c}{\textbf{Coh}} & \multicolumn{1}{c}{\textbf{Acc}} & \multicolumn{1}{c}{\textbf{Rich}} & \multicolumn{1}{c}{\textbf{Rel}}\\\hline
         MMPG+D & $4.42$& $2.75$& $3.15$ & $2.07$\\ 
         M-kplug & $4.48$& $3.21$& $3.18$ & $3.19$\\ 
         Oscar-GPT & $4.50$& $3.25$& $3.26$ & $3.17$\\
         ModICT(BART-L) & $\mathbf{4.63}$& $3.53$& $\mathbf{3.73}$ & $3.49$ \\ 
         ModICT(BLOOM-7B) & $4.54$& $\mathbf{3.61}$& $3.65$ & $\mathbf{3.53}$\\ 
         \hline
         Human & $4.81$& $3.72$& $4.33$ & $3.62$\\ 
         \hline 
    \end{tabular}
    \caption{Human evaluation results on the randomly selected sample set.}
        \label{tab:human_eva}
\end{table}

\begin{figure}[t]
    \centering
    \includegraphics[width=0.48\textwidth]{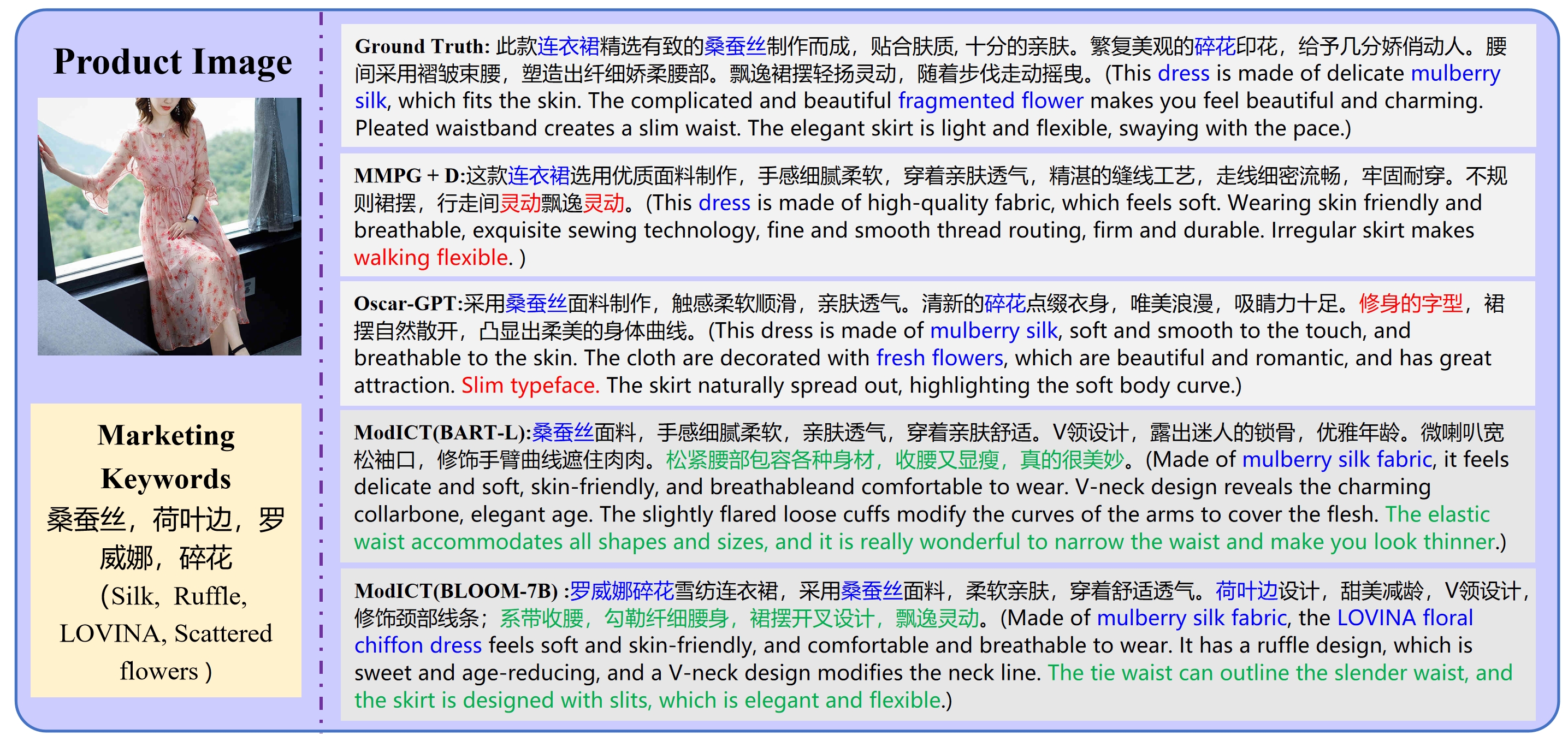}
    \caption{An illustration of descriptions generated by several models. The blue words represent keyphrases related to marketing keywords. Words in red show the inaccurate expression. The green-colored sentences are the eye-catching statements.}
    \label{fig:cases_exp}
\end{figure}

\subsection{Qualitative Analysis}
%The generated results are analyzed in terms of content accuracy, relevance to marketing keywords, and richness of rendering statements.  The sentences in green focus on the fine-grained features in the product image and improve the vividness of the product text (see the last two examples). 
\noindent\textbf{Case Study}. \begin{CJK}{UTF8}{gbsn}
We compare descriptions generated by various models in Figure~\ref{fig:cases_exp}. MMPG+D generates descriptions with universal characteristics but less relevance to marketing keywords. Oscar-GPT emphasizes keywords but can produce general and inaccurate statements similar to MMPG+D. The generated results of ModICT variants cover more aspects of the product and are more relevant to the marketing keywords, such as the words marked blue and green in the last generated example. In conclusion, ModICT variants could achieve superior performance in the accuracy, diversity, and vividness of the generated content.
\end{CJK}

\noindent\textbf{Human Evaluation.}
We conduct human evaluation on content accuracy (Acc), contextual coherence (Coh), richness (Rich), and relevance (Rel) to marketing keywords. Four master students rate ground truth and model-generated descriptions on 150 randomly selected testing samples. During scoring, model-generated and human-written descriptions are randomly shuffled and reviewed blindly. ModICT outperforms strong baselines in all aspects, particularly in content coherence and richness. However, there is still room for improvement in content richness compared to human-written descriptions. (See Table~\ref{tab:human_eva}).

%We further conduct an extensive human evaluation of the generated descriptions from four aspects: content accuracy (Acc), contextual coherence (Coh), richness (Rich), and relevance(Rel) to marketing keywords. We invite four master students who have taken part in related research works to rate the ground truth and description generated by models on $150$ randomly selected samples in the testing set. During scoring, model-generated and human-written descriptions are randomly shuffled and reviewed blindly. The score range in each evaluation aspect is set from one to five points, the higher the better. We report the average scores of several models in Table~\ref{tab:human_eva}. Results show that ModICT outperforms the strong baselines in all evaluation metrics from human perception. We also observe that the proposed method mainly improves the content coherence and richness. Compared to human-written descriptions, there is still a big gap in content richness.

\section{Conclusion}
In this work, we suggest a setting of E-commerce product description generation from images and marketing keywords, where the marketing keywords provide complementary information to the image. It could guide the generation of product descriptions to some extent by providing marketing keywords. To improve the accuracy and diversity of generated descriptions, we propose a simple and effective parameter-efficient multimodal in-context tuning approach, ModICT.

%via employing the powerful generation capability of frozen LLMs for multimodal generation tasks. The experimental results show that the proposed method can achieve significant improvement for various language models.

\section*{Ethics Statement}

The dataset included in our work is human-annotated E-commerce data that can be used for academic research, and we will release the preprocessing codes and data download source.

\section*{Acknowledge}
Thanks for the efforts from reviewers and action editors. This work is supported by grants: Natural Science Foundation of China (No. 62376067).

%\newpage
\section*{References}
\bibliographystyle{lrec-coling2024-natbib}
\bibliography{lrec-coling2024-example}

\begin{thebibliography}{60}
\expandafter\ifx\csname natexlab\endcsname\relax\def\natexlab#1{#1}\fi

\bibitem[{Brown et~al.(2020)Brown, Mann, Ryder, Subbiah, Kaplan, Dhariwal,
  Neelakantan, Shyam, Sastry, Askell et~al.}]{brown2020language_gpt3}
Tom Brown, Benjamin Mann, Nick Ryder, Melanie Subbiah, Jared~D Kaplan, Prafulla
  Dhariwal, Arvind Neelakantan, Pranav Shyam, Girish Sastry, Amanda Askell,
  et~al. 2020.
\newblock Language models are few-shot learners.
\newblock \emph{Advances in neural information processing systems},
  33:1877--1901.

\bibitem[{Chan et~al.(2019)Chan, Chen, Wang, Li, Zhang, Gai, Zhao, and
  Yan}]{chan2019stick_prod}
Zhangming Chan, Xiuying Chen, Yongliang Wang, Juntao Li, Zhiqiang Zhang, Kun
  Gai, Dongyan Zhao, and Rui Yan. 2019.
\newblock Stick to the facts: Learning towards a fidelity-oriented e-commerce
  product description generation.
\newblock In \emph{Proceedings of the 2019 Conference on Empirical Methods in
  Natural Language Processing and the 9th International Joint Conference on
  Natural Language Processing (EMNLP-IJCNLP)}, pages 4959--4968.

\bibitem[{Chen et~al.(2019)Chen, Lin, Zhang, Yang, Zhou, and Tang}]{know_pro}
Qibin Chen, Junyang Lin, Yichang Zhang, Hongxia Yang, Jingren Zhou, and Jie
  Tang. 2019.
\newblock Towards knowledge-based personalized product description generation
  in e-commerce.
\newblock In \emph{Proceedings of the 25th ACM SIGKDD International Conference
  on Knowledge Discovery \& Data Mining}, pages 3040--3050.

\bibitem[{Chen et~al.(2015)Chen, Fang, Lin, Vedantam, Gupta, Doll{\'a}r, and
  Zitnick}]{cococaption}
Xinlei Chen, Hao Fang, Tsung-Yi Lin, Ramakrishna Vedantam, Saurabh Gupta, Piotr
  Doll{\'a}r, and C~Lawrence Zitnick. 2015.
\newblock Microsoft coco captions: Data collection and evaluation server.
\newblock \emph{arXiv preprint arXiv:1504.00325}.

\bibitem[{Chen et~al.(2023{\natexlab{a}})Chen, Li, Zhao, Hu, and
  Zhang}]{chen2023temporal}
Ziyang Chen, Dongfang Li, Xiang Zhao, Baotian Hu, and Min Zhang.
  2023{\natexlab{a}}.
\newblock Temporal knowledge question answering via abstract reasoning
  induction.
\newblock \emph{arXiv preprint arXiv:2311.09149}.

\bibitem[{Chen et~al.(2023{\natexlab{b}})Chen, Liao, and
  Zhao}]{chen-etal-2023-multi}
Ziyang Chen, Jinzhi Liao, and Xiang Zhao. 2023{\natexlab{b}}.
\newblock \href {https://doi.org/10.18653/v1/2023.acl-long.637}
  {Multi-granularity temporal question answering over knowledge graphs}.
\newblock In \emph{Proceedings of the 61st Annual Meeting of the Association
  for Computational Linguistics (Volume 1: Long Papers)}, pages 11378--11392,
  Toronto, Canada. Association for Computational Linguistics.

\bibitem[{Cho et~al.(2021)Cho, Lei, Tan, and Bansal}]{pmlr-v139-cho21a}
Jaemin Cho, Jie Lei, Hao Tan, and Mohit Bansal. 2021.
\newblock \href {https://proceedings.mlr.press/v139/cho21a.html} {Unifying
  vision-and-language tasks via text generation}.
\newblock In \emph{Proceedings of the 38th International Conference on Machine
  Learning}, volume 139 of \emph{Proceedings of Machine Learning Research},
  pages 1931--1942. PMLR.

\bibitem[{Chowdhury et~al.(2021)Chowdhury, Bhowmik, Ravi, de~Melo, Razniewski,
  and Weikum}]{chowdhury2021exploiting}
Sreyasi~Nag Chowdhury, Rajarshi Bhowmik, Hareesh Ravi, Gerard de~Melo, Simon
  Razniewski, and Gerhard Weikum. 2021.
\newblock Exploiting image--text synergy for contextual image captioning.
\newblock In \emph{Proceedings of the Third Workshop on Beyond Vision and
  LANguage: inTEgrating Real-world kNowledge (LANTERN)}, pages 30--37.

\bibitem[{Desai and Johnson(2021)}]{Desai_2021_CVPR_caption}
Karan Desai and Justin Johnson. 2021.
\newblock Virtex: Learning visual representations from textual annotations.
\newblock In \emph{Proceedings of the IEEE/CVF Conference on Computer Vision
  and Pattern Recognition (CVPR)}, pages 11162--11173.

\bibitem[{Dosovitskiy et~al.(2020)Dosovitskiy, Beyer, Kolesnikov, Weissenborn,
  Zhai, Unterthiner, Dehghani, Minderer, Heigold, Gelly et~al.}]{vit}
Alexey Dosovitskiy, Lucas Beyer, Alexander Kolesnikov, Dirk Weissenborn,
  Xiaohua Zhai, Thomas Unterthiner, Mostafa Dehghani, Matthias Minderer, Georg
  Heigold, Sylvain Gelly, et~al. 2020.
\newblock An image is worth 16x16 words: Transformers for image recognition at
  scale.
\newblock \emph{arXiv preprint arXiv:2010.11929}.

\bibitem[{Du et~al.(2022)Du, Qian, Liu, Ding, Qiu, Yang, and
  Tang}]{du-etal-2022-glm}
Zhengxiao Du, Yujie Qian, Xiao Liu, Ming Ding, Jiezhong Qiu, Zhilin Yang, and
  Jie Tang. 2022.
\newblock \href {https://doi.org/10.18653/v1/2022.acl-long.26} {{GLM}: General
  language model pretraining with autoregressive blank infilling}.
\newblock In \emph{Proceedings of the 60th Annual Meeting of the Association
  for Computational Linguistics (Volume 1: Long Papers)}, pages 320--335,
  Dublin, Ireland. Association for Computational Linguistics.

\bibitem[{Elliott et~al.(2016)Elliott, Frank, Sima'an, and
  Specia}]{elliott2016multi30k}
Desmond Elliott, Stella Frank, Khalil Sima'an, and Lucia Specia. 2016.
\newblock Multi30k: Multilingual english-german image descriptions.
\newblock \emph{arXiv preprint arXiv:1605.00459}.

\bibitem[{Hao et~al.(2021)Hao, Guo, Wang, Liang, Yao, Wang, and
  Yu}]{deepdepict}
Shaoyang Hao, Bin Guo, Hao Wang, Yunji Liang, Lina Yao, Qianru Wang, and Zhiwen
  Yu. 2021.
\newblock \href {https://doi.org/10.1145/3446982} {Deepdepict: Enabling
  information rich, personalized product description generation with the deep
  multiple pointer generator network}.
\newblock \emph{ACM Trans. Knowl. Discov. Data}, 15(5).

\bibitem[{Hochreiter and Schmidhuber(1997)}]{hochreiter1997long}
Sepp Hochreiter and J{\"u}rgen Schmidhuber. 1997.
\newblock Long short-term memory.
\newblock \emph{Neural computation}, 9(8):1735--1780.

\bibitem[{Hu et~al.(2015)Hu, Chen, and Zhu}]{hu2015lcsts}
Baotian Hu, Qingcai Chen, and Fangze Zhu. 2015.
\newblock Lcsts: A large scale chinese short text summarization dataset.
\newblock \emph{arXiv preprint arXiv:1506.05865}.

\bibitem[{Huang et~al.(2016)Huang, Ferraro, Mostafazadeh, Misra, Agrawal,
  Devlin, Girshick, He, Kohli, Batra et~al.}]{huang2016visual}
Ting-Hao Huang, Francis Ferraro, Nasrin Mostafazadeh, Ishan Misra, Aishwarya
  Agrawal, Jacob Devlin, Ross Girshick, Xiaodong He, Pushmeet Kohli, Dhruv
  Batra, et~al. 2016.
\newblock Visual storytelling.
\newblock In \emph{Proceedings of the 2016 Conference of the North American
  Chapter of the Association for Computational Linguistics: Human Language
  Technologies}, pages 1233--1239.

\bibitem[{Jangra et~al.(2020)Jangra, Saha, Jatowt, and
  Hasanuzzaman}]{jangra2020multi_sum}
Anubhav Jangra, Sriparna Saha, Adam Jatowt, and Mohammad Hasanuzzaman. 2020.
\newblock Multi-modal summary generation using multi-objective optimization.
\newblock In \emph{Proceedings of the 43rd International ACM SIGIR Conference
  on Research and Development in Information Retrieval}, pages 1745--1748.

\bibitem[{Kayser et~al.(2021)Kayser, Camburu, Salewski, Emde, Do, Akata, and
  Lukasiewicz}]{kayser2021vil}
Maxime Kayser, Oana-Maria Camburu, Leonard Salewski, Cornelius Emde, Virginie
  Do, Zeynep Akata, and Thomas Lukasiewicz. 2021.
\newblock e-vil: A dataset and benchmark for natural language explanations in
  vision-language tasks.
\newblock In \emph{Proceedings of the IEEE/CVF International Conference on
  Computer Vision}, pages 1244--1254.

\bibitem[{Koh et~al.(2023)Koh, Salakhutdinov, and Fried}]{koh2023grounding}
Jing~Yu Koh, Ruslan Salakhutdinov, and Daniel Fried. 2023.
\newblock Grounding language models to images for multimodal generation.
\newblock \emph{arXiv preprint arXiv:2301.13823}.

\bibitem[{Li et~al.(2020{\natexlab{a}})Li, Yuan, Xu, Wu, He, and
  Zhou}]{li2020aspect_summ}
Haoran Li, Peng Yuan, Song Xu, Youzheng Wu, Xiaodong He, and Bowen Zhou.
  2020{\natexlab{a}}.
\newblock Aspect-aware multimodal summarization for chinese e-commerce
  products.
\newblock In \emph{Proceedings of the AAAI Conference on Artificial
  Intelligence}, volume~34, pages 8188--8195.

\bibitem[{Li et~al.(2020{\natexlab{b}})Li, Zhu, Zhang, He, and
  Zong}]{li2020multimodal_sentence-summariza}
Haoran Li, Junnan Zhu, Jiajun Zhang, Xiaodong He, and Chengqing Zong.
  2020{\natexlab{b}}.
\newblock Multimodal sentence summarization via multimodal selective encoding.
\newblock In \emph{Proceedings of the 28th International Conference on
  Computational Linguistics}, pages 5655--5667.

\bibitem[{Li et~al.(2020{\natexlab{c}})Li, Tang, Li, Xiao, Wu, Pu, and
  Zhuang}]{visual_story_tell}
Jiacheng Li, Siliang Tang, Juncheng Li, Jun Xiao, Fei Wu, Shiliang Pu, and
  Yueting Zhuang. 2020{\natexlab{c}}.
\newblock Topic adaptation and prototype encoding for few-shot visual
  storytelling.
\newblock In \emph{Proceedings of the 28th ACM International Conference on
  Multimedia}, pages 4208--4216.

\bibitem[{Li et~al.(2015)Li, Galley, Brockett, Gao, and
  Dolan}]{li2015diversity}
Jiwei Li, Michel Galley, Chris Brockett, Jianfeng Gao, and Bill Dolan. 2015.
\newblock A diversity-promoting objective function for neural conversation
  models.
\newblock \emph{arXiv preprint arXiv:1510.03055}.

\bibitem[{Li et~al.(2023{\natexlab{a}})Li, Li, Savarese, and Hoi}]{li2023blip}
Junnan Li, Dongxu Li, Silvio Savarese, and Steven Hoi. 2023{\natexlab{a}}.
\newblock Blip-2: Bootstrapping language-image pre-training with frozen image
  encoders and large language models.
\newblock \emph{arXiv preprint arXiv:2301.12597}.

\bibitem[{Li and Liang(2021)}]{li-liang-2021-prefix}
Xiang~Lisa Li and Percy Liang. 2021.
\newblock \href {https://aclanthology.org/2021.acl-long.353} {Prefix-tuning:
  Optimizing continuous prompts for generation}.
\newblock In \emph{Proceedings of the 59th Annual Meeting of the Association
  for Computational Linguistics and the 11th International Joint Conference on
  Natural Language Processing (Volume 1: Long Papers)}, pages 4582--4597,
  Online. Association for Computational Linguistics.

\bibitem[{Li et~al.(2020{\natexlab{d}})Li, Yin, Li, Hu, Zhang, Zhang, Wang, Hu,
  Dong, Wei, Choi, and Gao}]{li2020oscar}
Xiujun Li, Xi~Yin, Chunyuan Li, Xiaowei Hu, Pengchuan Zhang, Lei Zhang, Lijuan
  Wang, Houdong Hu, Li~Dong, Furu Wei, Yejin Choi, and Jianfeng Gao.
  2020{\natexlab{d}}.
\newblock Oscar: Object-semantics aligned pre-training for vision-language
  tasks.
\newblock \emph{ECCV 2020}.

\bibitem[{Li et~al.(2023{\natexlab{b}})Li, Hu, Ding, Ma, and
  Zhang}]{li-etal-2023-neural}
Yunxin Li, Baotian Hu, Yuxin Ding, Lin Ma, and Min Zhang. 2023{\natexlab{b}}.
\newblock \href {https://aclanthology.org/2023.acl-long.909} {A neural
  divide-and-conquer reasoning framework for image retrieval from
  linguistically complex text}.
\newblock In \emph{Proceedings of the 61st Annual Meeting of the Association
  for Computational Linguistics (Volume 1: Long Papers)}, pages 16464--16476,
  Toronto, Canada. Association for Computational Linguistics.

\bibitem[{Li et~al.(2023{\natexlab{c}})Li, Hu, Wang, Cao, and
  Zhang}]{li2023towards}
Yunxin Li, Baotian Hu, Wei Wang, Xiaochun Cao, and Min Zhang.
  2023{\natexlab{c}}.
\newblock Towards vision enhancing llms: Empowering multimodal knowledge
  storage and sharing in llms.
\newblock \emph{arXiv preprint arXiv:2311.15759}.

\bibitem[{Li et~al.(2023{\natexlab{d}})Li, Hu, Xinyu, Ding, Ma, and
  Zhang}]{li-etal-2023-multi-modal}
Yunxin Li, Baotian Hu, Chen Xinyu, Yuxin Ding, Lin Ma, and Min Zhang.
  2023{\natexlab{d}}.
\newblock \href {https://aclanthology.org/2023.acl-long.601} {A multi-modal
  context reasoning approach for conditional inference on joint textual and
  visual clues}.
\newblock In \emph{Proceedings of the 61st Annual Meeting of the Association
  for Computational Linguistics (Volume 1: Long Papers)}, pages 10757--10770,
  Toronto, Canada. Association for Computational Linguistics.

\bibitem[{Liang et~al.(2023{\natexlab{a}})Liang, Liu, Zhou, Tu, Wen, Yang,
  Dong, and Liu}]{Liangke_SymCLKG_TKDE}
Ke~Liang, Yue Liu, Sihang Zhou, Wenxuan Tu, Yi~Wen, Xihong Yang, Xiangjun Dong,
  and Xinwang Liu. 2023{\natexlab{a}}.
\newblock \href {https://doi.org/10.1109/TKDE.2023.3282989} {Knowledge graph
  contrastive learning based on relation-symmetrical structure}.
\newblock \emph{IEEE Transactions on Knowledge and Data Engineering}, pages
  1--12.

\bibitem[{Liang et~al.(2023{\natexlab{b}})Liang, Zhou, Liu, Meng, Liu, and
  Liu}]{liang2023structure}
Ke~Liang, Sihang Zhou, Yue Liu, Lingyuan Meng, Meng Liu, and Xinwang Liu.
  2023{\natexlab{b}}.
\newblock Structure guided multi-modal pre-trained transformer for knowledge
  graph reasoning.
\newblock \emph{arXiv preprint arXiv:2307.03591}.

\bibitem[{Lin(2004)}]{lin-2004-rouge}
Chin-Yew Lin. 2004.
\newblock {ROUGE}: A package for automatic evaluation of summaries.
\newblock In \emph{Text Summarization Branches Out}, pages 74--81, Barcelona,
  Spain. Association for Computational Linguistics.

\bibitem[{Lin and Ng(2019)}]{Lin_Ng_2019}
Hui Lin and Vincent Ng. 2019.
\newblock Abstractive summarization: A survey of the state of the art.
\newblock \emph{Proceedings of the AAAI Conference on Artificial Intelligence},
  33(01):9815--9822.

\bibitem[{Liu et~al.(2021)Liu, Ji, Fu, Du, Yang, and Tang}]{liu2021pv2}
Xiao Liu, Kaixuan Ji, Yicheng Fu, Zhengxiao Du, Zhilin Yang, and Jie Tang.
  2021.
\newblock P-tuning v2: Prompt tuning can be comparable to fine-tuning
  universally across scales and tasks.
\newblock \emph{arXiv preprint arXiv:2110.07602}.

\bibitem[{Long et~al.(2021)Long, Wang, and Li}]{long-etal-2021-generative}
Quanyu Long, Mingxuan Wang, and Lei Li. 2021.
\newblock \href {https://aclanthology.org/2021.naacl-main.457} {Generative
  imagination elevates machine translation}.
\newblock In \emph{Proceedings of the 2021 Conference of the North American
  Chapter of the Association for Computational Linguistics: Human Language
  Technologies}, pages 5738--5748, Online. Association for Computational
  Linguistics.

\bibitem[{Lu et~al.(2022)Lu, Zhu, Wang, Eckstein, and Wang}]{lu2022imagination}
Yujie Lu, Wanrong Zhu, Xin~Eric Wang, Miguel Eckstein, and William~Yang Wang.
  2022.
\newblock Imagination-augmented natural language understanding.
\newblock \emph{NACCL}.

\bibitem[{Novgorodov et~al.(2020)Novgorodov, Guy, Elad, and
  Radinsky}]{review-based}
Slava Novgorodov, Ido Guy, Guy Elad, and Kira Radinsky. 2020.
\newblock \href {https://doi.org/10.1145/3418202} {Descriptions from the
  customers: Comparative analysis of review-based product description
  generation methods}.
\newblock 20(4).

\bibitem[{Papineni et~al.(2002)Papineni, Roukos, Ward, and
  Zhu}]{papineni2002bleu}
Kishore Papineni, Salim Roukos, Todd Ward, and Wei-Jing Zhu. 2002.
\newblock Bleu: a method for automatic evaluation of machine translation.
\newblock In \emph{Proceedings of the 40th annual meeting of the Association
  for Computational Linguistics}, pages 311--318.

\bibitem[{Parida et~al.(2019)Parida, Bojar, and Dash}]{parida2019hindi}
Shantipriya Parida, Ond{\v{r}}ej Bojar, and Satya~Ranjan Dash. 2019.
\newblock Hindi visual genome: A dataset for multi-modal english to hindi
  machine translation.
\newblock \emph{Computaci{\'o}n y Sistemas}, 23(4):1499--1505.

\bibitem[{Parikh et~al.(2020)Parikh, Wang, Gehrmann, Faruqui, Dhingra, Yang,
  and Das}]{parikh-etal-2020-totto}
Ankur Parikh, Xuezhi Wang, Sebastian Gehrmann, Manaal Faruqui, Bhuwan Dhingra,
  Diyi Yang, and Dipanjan Das. 2020.
\newblock {ToTTo}: A controlled table-to-text generation dataset.
\newblock In \emph{Proceedings of the 2020 Conference on Empirical Methods in
  Natural Language Processing (EMNLP)}, pages 1173--1186, Online. Association
  for Computational Linguistics.

\bibitem[{Peng and Sollami(2022)}]{peng2022xfboost}
Xiangyu Peng and Michael Sollami. 2022.
\newblock Xfboost: Improving text generation with controllable decoders.
\newblock \emph{arXiv preprint arXiv:2202.08124}.

\bibitem[{Ren et~al.(2015)Ren, He, Girshick, and Sun}]{ren2015faster}
Shaoqing Ren, Kaiming He, Ross Girshick, and Jian Sun. 2015.
\newblock Faster r-cnn: Towards real-time object detection with region proposal
  networks.
\newblock \emph{Advances in neural information processing systems}, 28.

\bibitem[{Scao et~al.(2022)Scao, Fan, Akiki, Pavlick, Ili{\'c}, Hesslow,
  Castagn{\'e}, Luccioni, Yvon, Gall{\'e} et~al.}]{scao2022bloom}
Teven~Le Scao, Angela Fan, Christopher Akiki, Ellie Pavlick, Suzana Ili{\'c},
  Daniel Hesslow, Roman Castagn{\'e}, Alexandra~Sasha Luccioni, Fran{\c{c}}ois
  Yvon, Matthias Gall{\'e}, et~al. 2022.
\newblock Bloom: A 176b-parameter open-access multilingual language model.
\newblock \emph{arXiv preprint arXiv:2211.05100}.

\bibitem[{See et~al.(2017)See, Liu, and Manning}]{see-etal-2017-get}
Abigail See, Peter~J. Liu, and Christopher~D. Manning. 2017.
\newblock \href {https://doi.org/10.18653/v1/P17-1099} {Get to the point:
  Summarization with pointer-generator networks}.
\newblock In \emph{Proceedings of the 55th Annual Meeting of the Association
  for Computational Linguistics (Volume 1: Long Papers)}, pages 1073--1083,
  Vancouver, Canada. Association for Computational Linguistics.

\bibitem[{Shi et~al.(2019)Shi, Mao, Gimpel, and
  Livescu}]{shi-etal-2019-visually}
Haoyue Shi, Jiayuan Mao, Kevin Gimpel, and Karen Livescu. 2019.
\newblock \href {https://aclanthology.org/P19-1180} {Visually grounded neural
  syntax acquisition}.
\newblock In \emph{Proceedings of the 57th Annual Meeting of the Association
  for Computational Linguistics}, pages 1842--1861, Florence, Italy.
  Association for Computational Linguistics.

\bibitem[{Shi et~al.(2021)Shi, Liu, and Zhu}]{shi2021enhancing}
Zhan Shi, Hui Liu, and Xiaodan Zhu. 2021.
\newblock Enhancing descriptive image captioning with natural language
  inference.
\newblock In \emph{Proceedings of the 59th Annual Meeting of the Association
  for Computational Linguistics and the 11th International Joint Conference on
  Natural Language Processing (Volume 2: Short Papers)}, pages 269--277.

\bibitem[{Tang et~al.(2022)Tang, Wang, and Yao}]{tang2022dptdr}
Zhengyang Tang, Benyou Wang, and Ting Yao. 2022.
\newblock Dptdr: Deep prompt tuning for dense passage retrieval.
\newblock \emph{COLING}.

\bibitem[{Wang et~al.(2021)Wang, Tang, Yang, Bai, and Luo}]{wang2021improving}
Jing Wang, Jinhui Tang, Mingkun Yang, Xiang Bai, and Jiebo Luo. 2021.
\newblock Improving ocr-based image captioning by incorporating geometrical
  relationship.
\newblock In \emph{Proceedings of the IEEE/CVF Conference on Computer Vision
  and Pattern Recognition}, pages 1306--1315.

\bibitem[{Wu et~al.(2022)Wu, Zhao, Hu, Minervini, Stenetorp, and
  Riedel}]{wu2022efficient}
Yuxiang Wu, Yu~Zhao, Baotian Hu, Pasquale Minervini, Pontus Stenetorp, and
  Sebastian Riedel. 2022.
\newblock An efficient memory-augmented transformer for knowledge-intensive nlp
  tasks.
\newblock \emph{arXiv preprint arXiv:2210.16773}.

\bibitem[{Xu et~al.(2021)Xu, Li, Yuan, Wang, Wu, He, Liu, and
  Zhou}]{xu-etal-2021-k-plug}
Song Xu, Haoran Li, Peng Yuan, Yujia Wang, Youzheng Wu, Xiaodong He, Ying Liu,
  and Bowen Zhou. 2021.
\newblock K-{PLUG}: Knowledge-injected pre-trained language model for natural
  language understanding and generation in {E}-commerce.
\newblock In \emph{Findings of the Association for Computational Linguistics:
  EMNLP 2021}, pages 1--17, Punta Cana, Dominican Republic. Association for
  Computational Linguistics.

\bibitem[{Yang et~al.(2022)Yang, Pan, Lin, Men, Zhang, Zhou, and
  Zhou}]{chinese-clip}
An~Yang, Junshu Pan, Junyang Lin, Rui Men, Yichang Zhang, Jingren Zhou, and
  Chang Zhou. 2022.
\newblock Chinese clip: Contrastive vision-language pretraining in chinese.
\newblock \emph{arXiv preprint arXiv:2211.01335}.

\bibitem[{Yang et~al.(2021)Yang, Wu, Hu, Xu, Wang, and Li}]{yang2021open}
Ze~Yang, Wei Wu, Huang Hu, Can Xu, Wei Wang, and Zhoujun Li. 2021.
\newblock Open domain dialogue generation with latent images.
\newblock In \emph{Proceedings of the AAAI Conference on Artificial
  Intelligence}, volume~35, pages 14239--14247.

\bibitem[{Yuan et~al.(2020)Yuan, Li, Xu, Wu, He, and
  Zhou}]{yuan-etal-2020-faithfulness}
Peng Yuan, Haoran Li, Song Xu, Youzheng Wu, Xiaodong He, and Bowen Zhou. 2020.
\newblock On the faithfulness for {E}-commerce product summarization.
\newblock In \emph{Proceedings of the 28th International Conference on
  Computational Linguistics}, pages 5712--5717, Barcelona, Spain (Online).
  International Committee on Computational Linguistics.

\bibitem[{Zhan et~al.(2021)Zhan, Zhang, Chen, Shen, Ding, Bao, Yan, and
  Lan}]{zhan2021probing}
Haolan Zhan, Hainan Zhang, Hongshen Chen, Lei Shen, Zhuoye Ding, Yongjun Bao,
  Weipeng Yan, and Yanyan Lan. 2021.
\newblock Probing product description generation via posterior distillation.
\newblock In \emph{Proceedings of the AAAI Conference on Artificial
  Intelligence}, volume~35, pages 14301--14309.

\bibitem[{Zhang et~al.(2019)Zhang, Zhang, Huo, and Ren}]{pattern_product}
Tao Zhang, Jin Zhang, Chengfu Huo, and Weijun Ren. 2019.
\newblock \href {https://doi.org/10.1145/3308558.3313407} {Automatic generation
  of pattern-controlled product description in e-commerce}.
\newblock In \emph{The World Wide Web Conference}, WWW '19, page 2355–2365,
  New York, NY, USA. Association for Computing Machinery.

\bibitem[{Zhang et~al.(2020)Zhang, Kishore, Wu, Weinberger, and
  Artzi}]{zhang2019bertscore}
Tianyi Zhang, Varsha Kishore, Felix Wu, Kilian~Q Weinberger, and Yoav Artzi.
  2020.
\newblock Bertscore: Evaluating text generation with bert.
\newblock \emph{International Conference on Learning Representations (ICLR)}.

\bibitem[{Zhang et~al.(2021)Zhang, Zhang, Chen, Guo, Hua, Wang, and
  Zhou}]{zhang2021mengzi}
Zhuosheng Zhang, Hanqing Zhang, Keming Chen, Yuhang Guo, Jingyun Hua, Yulong
  Wang, and Ming Zhou. 2021.
\newblock Mengzi: Towards lightweight yet ingenious pre-trained models for
  chinese.
\newblock \emph{arXiv preprint arXiv:2110.06696}.

\bibitem[{Zhu et~al.(2018)Zhu, Li, Liu, Zhou, Zhang, and Zong}]{zhu2018msmo}
Junnan Zhu, Haoran Li, Tianshang Liu, Yu~Zhou, Jiajun Zhang, and Chengqing
  Zong. 2018.
\newblock Msmo: Multimodal summarization with multimodal output.
\newblock In \emph{Proceedings of the 2018 conference on empirical methods in
  natural language processing}, pages 4154--4164.

\bibitem[{Zhu et~al.(2020)Zhu, Zhou, Zhang, Li, Zong, and Li}]{product_summ}
Junnan Zhu, Yu~Zhou, Jiajun Zhang, Haoran Li, Chengqing Zong, and Changliang
  Li. 2020.
\newblock \href {https://doi.org/10.1609/aaai.v34i05.6525} {Multimodal
  summarization with guidance of multimodal reference}.
\newblock \emph{Proceedings of the AAAI Conference on Artificial Intelligence},
  34(05):9749--9756.

\bibitem[{Zhu et~al.(2022)Zhu, Yan, Lu, Xu, Wang, Eckstein, and
  Wang}]{zhu2022visualize}
Wanrong Zhu, An~Yan, Yujie Lu, Wenda Xu, Xin~Eric Wang, Miguel Eckstein, and
  William~Yang Wang. 2022.
\newblock Visualize before you write: Imagination-guided open-ended text
  generation.
\newblock \emph{arXiv preprint arXiv:2210.03765}.

\end{thebibliography}

%\section{Language Resource References}
%\label{lr:ref}
%\bibliographystylelanguageresource{lrec-coling2024-natbib}
%\bibliographylanguageresource{languageresource}

\end{document}